\documentclass{Interspeech}
\usepackage{svg}
\newcommand\anonymize[1]{[ANONYMIZED]}

\interspeechcameraready

\title{Modeling Probabilistic Reduction using Information Theory and Naive Discriminative Learning}

\author[affiliation={1}]{Anna}{Stein}
\author[affiliation={2,3}]{Kevin}{Tang}

\affiliation{Institute of Linguistics}{Faculty of Arts and Humanities, HHU Düsseldorf}{Germany}
\affiliation{Department of English Language and Linguistics}{Faculty of Arts and Humanities, HHU Düsseldorf}{Germany}
\affiliation{Department of Linguistics}{University of Florida}{USA}
\email{anna.stein@hhu.de, kevin.tang@hhu.de}

\usepackage{comment}

\begin{document}

\maketitle

\begin{abstract}
This study compares probabilistic predictors based on information theory with Naive Discriminative Learning (NDL) predictors in modeling acoustic word duration, focusing on probabilistic reduction. We examine three models using the Buckeye corpus: one with NDL-derived predictors using information-theoretic formulas, one with traditional NDL predictors, and one with N-gram probabilistic predictors. Results show that the N-gram model outperforms both NDL models, challenging the assumption that NDL is more effective due to its cognitive motivation. However, incorporating information-theoretic formulas into NDL improves model performance over the traditional model. This research highlights a) the need to incorporate not only frequency and contextual predictability but also average contextual predictability, and b) the importance of combining information-theoretic metrics of predictability and information derived from discriminative learning in modeling acoustic reduction. 
\end{abstract}

\section{Introduction} 



A consistent influence on the acoustic duration of a word is its predictability \cite{gahl2008time,Zipf1929,aylett2004smooth}. Predictability is typically quantified using information theoretic/probability-based predictors such as frequency, contextual predictability, and informativity \cite{seyfarth2014word, CohenPriva2018}. However, recent research on acoustic duration has seen the emergence of a new set of predictors derived from discriminative learning \cite{tomaschek2021phonetic, SteinTang2024_Interspeech}\footnote{\url{https://osf.io/preprints/psyarxiv/jc97w/}}. While both the types of predictors can effectively model probabilistic reduction, they differ in their theoretical assumptions and computation.

Probability measures are based on information theory \cite{shannon1948mathematical}, specifically the concept of surprisal \cite{gibson2019efficiency}. The main idea is that speakers try to keep a steady flow of information in their speech, as suggested, for example, by the ``Smooth Signal Redundancy hypothesis'' \cite{aylett2004smooth}. Research on probabilistic reduction focuses on three main metrics: frequency, contextual predictability, and informativity. The inverse relationship of frequency and duration is well established: words with a high frequency are produced with shortened acoustic duration \cite{Zipf1929,gahl2008time,bell2003effects}. Contextual predictability describes the predictability of a word in a given context \cite{bell2003effects, seyfarth2014word}. This idea is linked to surprisal in, for instance, \cite{gibson2019efficiency, wilcox2023}. There can also be words that, while frequent overall, are much less predictable in the individual context in which they appear. Consider the example from \cite[p.141]{seyfarth2014word} ``the word `current' usually occurs in the context of `current events' or the `current situation', and is therefore usually predictable in context. On the other hand, the word `nowadays' has roughly the same log-frequency overall as `current', but 'nowadays' occurs in a wide variety of contexts.''. This information is captured by informativity, the average predictability of a word. Studies have used these metrics to model speech in various languages, including English \cite{seyfarth2014word, CohenPriva2018}, Chinese \cite{tang2021prosody}, Kaqchikel Mayan \cite{tang2018contextual}, and Japanese \cite{hashimoto2021}, typically relying on N-gram or large language models. 

Conversely, studies using Naive Discriminative Learning (NDL) derive predictors from a model that aims to model implicit, low-level learning based on principles from Pavlovian conditioning. NDL practitioners do not use the information-theoretic formulation of surprisal and instead opt for a combination of summing row and column values, depending on the training data and research question \cite{chuang2021discriminative}. This may be because an NDL model outputs a weight matrix that is not on the probability scale, and information theoretic measures are formulated for probabilities. Previous studies using NDL to model acoustic duration have tried to compare the probabilistic metrics and the NDL metrics but either ended up excluding them due to correlation \cite{tomaschek2021phonetic} or do not use the full range of predictors\footnote{\url{https://osf.io/preprints/psyarxiv/jc97w/}}. Nonetheless, these studies effectively model duration using NDL predictors, even without probability-based metrics \cite{SteinTang2024_Interspeech}. 

We bring both probabilistic metrics and NDL-based metrics together in our study. We use the cognitively motivated, non-static, Pavlovian learning model (NDL) and the information-theoretically motivated formulas for calculating metrics. Our aim is to investigate whether this approach leads to better results when modeling probabilistic reduction and investigate the relationship between the two sets of predictors. Specifically, we aim to answer the following questions: \textbf{1) What is the relationship between the probabilistic NDL metrics, the traditional NDL metrics, and the N-gram model-derived probabilistic metrics?} \textbf{2) Do metrics derived from an NDL model calculated with the information-theoretic formula produce a better model than one derived from an N-gram model or one with traditional NDL measures?}

To address these questions, we train three models using the same corpus: one based on NDL-based predictors, calculated using the information-theoretic formula; one based on NDL predictors calculated the ``traditional'' way according to previous research; and another on probabilistic predictors from a language model. The dependent variable and other predictors for our regression analysis are derived from tokens in the Buckeye corpus, which contains interview speech data from 40 American speakers. A linear mixed effects model is used to compare the three models, one for each set of predictors. 

Our study\footnote{Our code and data is available \url{https://doi.org/10.5281/zenodo.15525666}.} presents several new elements to the topic of probabilistic reduction: Unlike previous studies\footnote{\url{https://osf.io/preprints/psyarxiv/jc97w/}}, we include all three main predictors associated with probabilistic reduction when comparing them to the NDL-based predictors. Additionally, we present a new set of predictors derived using the information-theoretic formulas used for calculating contextual predictability and informativity. We reveal that models using N-gram-based probabilistic predictors outperform both traditional and information-theoretic NDL models in predicting word duration. However, incorporating information-theoretic formulas into NDL-based predictors improves their performance compared to the traditional NDL approach. Our findings 1) provide NDL practitioners with a new set of measures that are theoretically motivated, 2) challenge the notion that cognitively motivated models like NDL are inherently superior 3) provide a better understanding of the factors influencing acoustic duration, in particular in the context of probabilistic reduction. 

\section{Materials and Methods}
\subsection{Acoustic corpus}\label{sec:acoustic-corpus}
The Buckeye corpus of conversational speech \cite{pitt2005buckeye} includes interviews with 40 speakers from Ohio, USA. We used the \texttt{buckeye} package\footnote{\url{https://github.com/scjs/buckeye}} (\texttt{v1.3}) to extract tokens and their metadata, focusing only on words and excluding pauses. Following \cite{seyfarth2014word}, we removed tokens adjacent to pauses, those at the start or end of utterances, and tokens with negative durations. An utterance is defined as a conversational turn marked by pauses. We also added word boundary tokens (\texttt{<s>} and \texttt{</s>}) and removed tokens with durations over \SI{10}{\second}, as well as words not present in the training corpus.

\subsection{Training corpus}
To examine the effect of word probabilities, a speech-like written corpus of English was used to estimate word frequency, contextual predictability, and informativity. The corpus consists of \num{446612} subtitle files from TV/film in English and was sourced from OpenSubtitle (\texttt{v2018})\footnote{\url{http://www.opensubtitles.org/}} \cite{Tiedemann2009}. Subtitle texts were chosen as a genre because i) previous work has shown that probabilistic estimates derived from subtitle texts consistently outperform those from non-speech-like genres in explaining behavioral data \cite{New2007}; ii) subtitle texts are a better genre-of-speech match with the conversational speech compared to written texts; iii) informativity computed using a large corpus will reduce the chance of frequency falsely capturing effects that should be attributed to informativity \cite{CohenPriva2018}.

The corpus was preprocessed by removing XML tags and non-ASCII characters, and converting numbers with the \texttt{inflect} package\footnote{\url{https://github.com/jaraco/inflect/}} (\texttt{v7.4.0}). We excluded 653 non-English files using \texttt{fast\_langdetect}\footnote{\url{https://github.com/LlmKira/fast-langdetect}} (\texttt{v0.2.5}). To address subtitle formatting issues, we merged lines starting with a lowercase letter into the previous line unless the preceding line ended with a question mark, exclamation point, or period. Sentence boundaries were marked with \texttt{<s>} and \texttt{</s>}, and any separated contractions were merged. 

We randomly selected a subset of \num{10000} files ($\approx$72 million words) from the corpus for model training. Tokens that appears in less than six files in the entire corpus, not just the subset, were replaced with an \texttt{<unk>} token.

\subsection{N-gram Language Model}
Two bigram language models were created using the training corpus: one predicts each word based on the previous word, and the other on the following word. Model construction was carried out using the MIT Language Modeling (MITLM) toolkit\footnote{\url{http://code.google.com/p/mitlm/}} \texttt{v0.4.2} with the modified Kneser-Ney smoothing using default parameters. Training was done locally on a system with 22GB RAM and 16 cores, taking about a minute. The three-word probability measures are detailed below.

\textbf{Word frequency} is the total number of times a word appears in the lexical corpus. Following \cite{CohenPriva2018}, a maximum likelihood estimation of the type probability was derived by dividing the frequency of the token by the sum of all frequencies in the Buckeye corpus and $\log_2$ transforming it.

\textbf{Contextual predictability} is the conditional probability of a word given its context (see \ref{eq:cp}), where $c$ is the context, which is implemented as the preceding or following word in an utterance, and $w$ is the target word.

\begin{align}
    log2(Pr(W = w \mid C = c))
    \label{eq:cp}
\end{align}

\textbf{Informativity} is the negative log average contextual predictability of a word in every context in which it appears in, weighted by the contextual predictability of the contexts given the word, as formulated in (\ref{eq:informativity}).

\begin{align}
    - \sum_{c} Pr(C = c \mid W = w) \log_2 Pr(W = w \mid C = c)
    \label{eq:informativity}
\end{align}

\subsection{Naive Discriminative Learning (NDL)}
Naive Discriminative Learning (NDL) implements a form of discriminative learning based on Pavlovian conditioning \cite{rescorlawagner1972}. It is naive in the sense that it does not assume any kind of form of representation. Instead, input is given via learning events containing cues and outcomes. In a given learning event, cues can be either present or absent. This binary representation determines the update of the cues' connection to the outcome in the learning event by either increasing or decreasing it. Each time a cue that has previously occurred with an outcome is not present in a learning event, its weight to that outcome is increased. This results in a ``cue competition'' where each cue tries to be the most informative for a given outcome (see \cite{chuang2021discriminative} for an in-depth mathematical overview of the learning algorithm).

\subsection{Model training and feature extraction} \label{sec:feature-extraction}
We trained two models using \texttt{pyndl} \cite{sering2022pyndl}, with one learning event comprising a target's preceding or following word as a cue and the target word as an outcome. We did not split the data into training, development, and testing sets since inference was performed using the Buckeye corpus, and our focus was not on improving model accuracy. The models were trained in batches on the HPC system at HHU with 32GB RAM and 12 CPU cores, taking about eight hours each (parameters used: learning rate $\alpha = 0.001$, maximum connection strength $\lambda = 1.0$, and weight adjustments $\beta_i = 0.1$, $\beta_j = 0.1$, as is standard practice for NDL modeling \cite{tomaschek2021phonetic}).

\textbf{NDL-based contextual predictability and informativity}
This measure aims to align with the contextual predictability formula using NDL weights. To avoid numerical under-/overflow, we first balance the outcome weights by subtracting the highest value in the column \cite{Goodfellow-et-al-2016}. We then convert these weights into probabilities using the \textit{log\_softmax} function from the \textit{scipy} package \cite{2020SciPy-NMeth}. Informativity is calculated using the same formula (\ref{eq:informativity}) as the N-gram model.

\textbf{Prior}
The Prior of a word is calculated by summing the absolute values of its column in the weight matrix, serving as a measure of pre-existing availability. Previous studies show a correlation between Prior and frequency \cite{milin2017discrimination, tomaschek2021phonetic}. For our regression analysis, we compute the Prior for each word type in both the preceding and following models.

\textbf{Activation} Activation is a measure used in NDL studies, representing the "support" of a cue for a specific outcome. It is obtained by selecting the weight cell of a cue-outcome pair. We compute this for each word in the previous and following context models.
\section{Analysis}
\subsection{Correlation Analysis}
We carry out a correlation analysis to explore how probability-based from the NDL and N-gram predictors, traditional NDL predictors, and duration relate to each other. First, we want to establish whether our results match previous research on probabilistic measures. We also look at the correlation between probabilistic measures and the new probabilistic NDL measures; if we find a strong correlation, we may need to adjust our regression analysis. Since the new probabilistic NDL predictors have not been used before, we examine how they correlate with other variables and with existing NDL predictors to see if they add new insights. The results are shown in Table \ref{tab:correlations} using the Pearson correlation coefficient.

\begin{table*}[t]
 \caption{Correlation values of all predictors and duration.}
 \label{tab:correlations}
 \centering
\begin{tabular}{lcccccccccccc}
\toprule 
 & \textbf{1.} & \textbf{2.} & \textbf{3.} & \textbf{4.} & \textbf{5.} & \textbf{6.} & \textbf{7.} & \textbf{8.}& \textbf{9.} & \textbf{10.} & \textbf{11.} & \textbf{12.}\\
 \midrule 
\textbf{1.} N-gram cp prev. & 1 &&&&&&&&&&\\
\textbf{2.} N-gram cp foll. & 0.39 & 1 & &&&&&&&&\\
\textbf{3.} N-gram inf foll. & -0.06 & 0.16 & 1 & &&&&&&&&\\
\textbf{4.} N-gram inf prev. &0.48&0.01&-0.01&1  &&&&&&&&\\
\textbf{5.} Ndl inf foll. &-0.04&-0.14&\textbf{0.68}&-0.01&1&&&&&&&\\
\textbf{6.} Ndl inf prev. &0.50&-0.04&-0.01&\textbf{0.95}&-0.01&1&&&&&&\\
\textbf{7.} Ndl cp foll. &0.12&\textbf{0.34}&-0.06&0.01&-0.09&0.00&1&&&&&\\
\textbf{8.} Ndl cp prev. &\textbf{0.24}&0.1&-0.01&-0.01&-0.01&-0.03&0.11&1&&&&\\
\textbf{9.} Ndl prior foll. &0.3&0.37&-0.08&-0.05&-0.03&-0.11&0.14&0.09&1&&&\\
\textbf{10.} Ndl prior prev. &0.20&0.18&-0.15&-0.02&-0.06&-0.05&0.08&-0.02&\textbf{0.66}&1&&\\
\textbf{11.} Frequency &0.59&0.63&-0.11&0.02&-0.05&-0.04&0.17&0.1&0.47&0.5&1&\\
\textbf{12.} Duration &\textbf{-0.35}&\textbf{-0.4}&\textbf{0.13}&\textbf{0.011}&\textbf{0.09}&\textbf{0.06}&\textbf{-0.14}&\textbf{-0.08}&\textbf{-0.56}&\textbf{-0.45}&\textbf{-0.67}&1\\
\bottomrule
\end{tabular}
\end{table*}

\subsection{Correlation Results}
We observe that predictors of contextual predictability and frequency show a negative correlation with duration, while informativity predictors exhibit a positive correlation, consistent with previous research \cite{seyfarth2014word} (Table \ref{tab:correlations}). Most probabilistic measures and NDL predictors do not display near-perfect correlations, with contextual predictability predictors (1., 2., 7., 8.) ranging from $r$ = 0.1 to 0.34. Higher correlations are found among informativity predictors, especially within the same context direction (previous $r$ = 0.95, following $r$ = 0.68). The moderate correlations indicate that while these predictors share similarities, they also maintain distinct characteristics, supporting our choice for model comparison.


\subsection{Regression Analysis}
\subsubsection{Fixed effect variables: Control variables}
A number of control variables were included as they have been found to influence word duration independently (e.g., \cite{tomaschek2021phonetic}). Word length, was included in form of the \texttt{number of segments and syllables} of a token. \texttt{DeepPhonemizer}\footnote{\url{https://github.com/as-ideas/DeepPhonemizer}} (\texttt{v 0.0.17}) was used to generate segment transcriptions which were then syllabified using a modified version of the \texttt{P2K toolkit}\footnote{\url{https://sourceforge.net/projects/p2tk/}}. The Buckeye corpus provides metadata for binary \texttt{speaker and interviewer gender} (reference level: female), speaker age (`young': $<$ 40 and `old': $>$ 40) \cite{kiesling2006variation} (both contrast coded (-0.5, 0.5)). The corpus also provides the syntactic category for each token. We excluded all grammatical categories to only have lexical categories (adjectives, verbs, nouns, and adverbs). We coded the \texttt{part-of-speech} variable with the target encoding scheme \cite{MicciBarreca2001}, which takes the mean of the dependent variable (duration) for each category producing a single continuous variable. We calculated \texttt{speech rate} as the number of syllables per utterance (as defined in \ref{sec:acoustic-corpus}) divided by the total duration of the utterance.

\subsubsection{Statistical procedures}
To compare how our two sets of predictors predict word duration, we used a linear mixed-effects model implemented in \textit{R} \cite{r} with the \textit{lme4} package \cite{lme4}. We included both Speaker and Word as random effects. To compare the relative effect sizes ($\beta$) of the predictors, we standardized all continuous variables using \textit{z}-transformation. Additionally, the response variable (duration in milliseconds), was log-transformed to the base 10.

\subsection{Regression Results}
We compare three models: one using probabilistic predictors from the N-gram model, another with traditional NDL predictors, and a third with the new probability-like NDL predictors. The results presented in Table \ref{tab:effect-sizes} include effect sizes ($\beta$), p-values, Aikake Information Criterion (AIC), and Bayesian Information Criterion (BIC) for each model. All control variables were significant except for speaker gender, interviewer gender, and speaker age. The listed models all show improvements over the baseline model, which includes only the control variables (AIC: -65298, BIC: -65202).

The N-gram model shows the best performance, with the lowest AIC and BIC scores, followed by the probabilistic NDL model. The traditional NDL model has the worst performance. All predictors, except contextual predictability from the probabilistic NDL model, and the N-gram model, are significant. Frequency or its proxy in the NDL models has the largest effect size. Notably, informativity exceeds the effect size of contextual predictability.

We also observe suppressor effects in the N-gram model, where contextual predictability and informativity predictors show opposite signs compared to the correlation analysis.

\begin{table}[th]
 \caption{Fixed effect summary for probabilistic NDL predictors, the n-gram predictors and the traditional NDL predictors. Bold values are statistically significant (p-value $<$ 0.05).}
 \label{tab:effect-sizes}
 \centering
 \begin{tabular}{lccc}
 \toprule
 \multicolumn{1}{c}{} & 
 \multicolumn{1}{c}{\textbf{Prob. NDL}} &
 \multicolumn{1}{c}{\textbf{N-gram}} &
 \multicolumn{1}{c}{\textbf{Trad. NDL}} \\
 
 \multicolumn{1}{c}{} & 
 \multicolumn{1}{c}{\textbf{$\beta$ ($10^{-2}$)}} &
 \multicolumn{1}{c}{\textbf{$\beta$ ($10^{-2}$)}} &
 \multicolumn{1}{c}{\textbf{$\beta$ ($10^{-2}$)}} \\
 \midrule
 Prior prev.              & \textbf{-6.85}    & -                & \textbf{-6.73}\\
 Prior foll.              & \textbf{-1.32}    & -                & \textbf{-1.34}\\
 Frequency                & -                 & \textbf{-3.68}   & -\\
 \midrule
 Act. prev.               & -                 & -                & \textbf{-0.39}\\
 Act. foll.               & -                 & -                & \textbf{-0.16}\\
 Cont. pred. prev.        & \textbf{-0.38}    & \textbf{0.58}    & -\\
 Cont. pred. foll.        & -0.11             & 0.14             & -\\
 \midrule
 Inf. prev.               & \textbf{0.30}     & \textbf{-0.24}    & -\\
 Inf. foll.               & \textbf{1.07}    & \textbf{1.06}    & -\\
 \midrule
 \midrule
 AIC                      & -65939            &  -66028          & -65543 \\
 BIC                      & -65790            &  -65888          & -65412\\
 \bottomrule
 \end{tabular} 
\end{table}

\section{Discussion}
When comparing the probabilistic NDL model and the N-gram probabilistic model, we found that the N-gram probabilistic predictors produced a better model, even when both sets of measures were calculated using the same formula. Noticeably, the main difference between the approaches was the training algorithm.

In our study, NDL was found to be less effective than the N-gram model, which contradicts the findings by \footnote{\url{https://osf.io/preprints/psyarxiv/jc97w/}}, who reported improved results with NDL predictors compared to frequency-based ones. It, therefore, cannot be assumed that NDL, compared to an N-gram model, produces inherently better results because it tries to incorporate human learning mechanisms. Conversely, our findings confirm that probabilistic metrics enhance the performance of the NDL model when compared to traditional NDL measures. Our main findings indicate that a model using probabilistic metrics outperforms one that uses traditional NDL metrics. This suggests that NDL practitioners should adopt the information-theoretically motivated formula.

One limitation of the probabilistic model is its vulnerability to various suppressor effects, which do not impact the NDL model. In the NDL model, all effects correspond with the sign shown in the correlation analysis. The suppressor effect likely arises from the high correlation between frequency and contextual predictability, resulting in reversed signs for contextual predictability.

Regarding the comparison between the model using previous context and the model using following context, we see that measures derived from the previous context are generally more significant than those based on the following context in our model. This finding does not align with the results from \cite{seyfarth2014word} and \cite{hashimoto2021probabilistic}, but is consistent with the ones by \cite{tang2021prosody} for Chinese.

Our correlation analysis supports the findings of \cite{tomaschek2021phonetic}, which demonstrate that Prior correlates with frequency. However, frequency/Prior cannot simply be replaced by informativity. Our analysis suggests that frequency/Prior serves as a more effective predictor than informativity, which is in turn more reliable than contextual predictability. Based on our correlation analysis, one might expect probabilistic predictors to outperform the probabilistic NDL variables if not for the influence of the Prior measure. Specifically, the Prior measure seems to be responsible for the reduction effect observed in our study. This stands in contrast to the conclusions drawn by \cite{seyfarth2014word} and \cite{CohenPriva2018}, which identified informativity as the principal influencing factor.

\section{Limitations} 
First of all, using a broader range of probabilistic predictors could enhance the robustness of our comparisons. Second, our focus on a context window size of one may limit the model's ability to capture longer-range dependencies; thus, exploring larger context windows might lead to improved performance. However, \cite{hao-etal-2022-ji} has shown that N-gram models can pick up on the same results as a GPT-2 model. Additionally, we excluded activation diversity, which is a common metric in NDL studies, because it is conceptually similar to neighborhood density, and we wish to restrict our comparison to only probabilistic predictors. Our findings regarding directionality differ from those of \cite{seyfarth2014word}, likely due to the differing control variables used; therefore, further exploration of these directional effects is warranted.

\section{Conclusion}
This study compared probabilistic and Naive Discriminative Learning (NDL) predictors for modeling acoustic word duration. We established the relative independence of three sets of metrics that can be used to capture probabilistic reduction, which suggests that the choice of training algorithm and the mathematical formulation of predictability are both important factors. We found that N-gram-based probabilistic predictors outperformed both NDL models, challenging the idea that NDL's cognitive motivation makes it superior. However, incorporating information-theoretic formulas into NDL improved its performance over traditional NDL predictors. These findings emphasize the importance of information-theoretic metrics, regardless of the training algorithm (n-gram language model or discriminative learning). They also underline the importance of combining probabilistic and discriminative learning metrics, suggesting that frequency and contextual predictability are crucial factors in modeling probabilistic reduction.




\section{Acknowledgements}
CRediT author statement: Conceptualization, Investigation, Methodology, Formal Analysis, Writing – original draft/review/editing: AS, KT; Data Curation, Project Administration, Software, Validation: AS; Resources, Supervision: KT

We thank Fabian Tomaschek for helping us train the NDL model and acknowledge the support of the central HPC system ``HILBERT'' at Heinrich Heine University Düsseldorf. We would also like to thank those who contributed to this work: Anh Kim Nguyen, Eoin O'Reilly, Jennifer Keller, and Chris Geissler. Finally, we thank the ``scholarships for the advancement of women'' who funded the trip for one author.


\bibliographystyle{IEEEtran}
\bibliography{mybib}

\end{document}